\documentclass{article}

\usepackage{arxiv}
\usepackage[breaklinks=true,colorlinks,citecolor=blue,linkcolor=blue,urlcolor=blue]{hyperref}
\usepackage{graphicx}
\usepackage{enumitem}
\usepackage{amsmath}
\usepackage{physics}
\def\b{{\mathbf b}}
\def\x{{\mathbf x}}
\def\t{{\mathbf t}}
\def\v{{\mathbf v}}
\def\p{{\mathbf p}}

\title{CDRNet: Accurate Cup-to-Disc Ratio Measurement with Tight Bounding Box Supervision in Fundus Photography Using Deep Learning}

\date{} 					

\author{ {Juan~Wang} \\
	Delta Micro Technology, Inc.\\
	23421 S. Pointe Dr.\\
	Laguna Hills, CA 92653 \\
	\texttt{wangjuan313@gmail.com} \\
	\And
	{Bin~Xia} \\
	Shenzhen SiBright Co., Ltd.\\
	Tinwe Industrial Park, No. 6 Liufang Rd.\\
	Shenzhen, Guangdong 518052 \\
	\texttt{b.xia@sibionics.com} \\
}

\renewcommand{\headeright}{}
\renewcommand{\undertitle}{}
\renewcommand{\shorttitle}{CDRNet}

\hypersetup{
pdftitle={CDRNet: Accurate Cup-to-Disc Ratio Measurement with Tight Bounding Box Supervision in Fundus Photography Using Deep Learning},
pdfsubject={cs.AI, cs.CV},
pdfauthor={Juan~Wang, Bin~Xia},
pdfkeywords={Cup-to-disc ratio (CDR), Bounding box tightness prior, Class-specific bounding-box regression, Weakly supervised image segmentation, Fundus photography},
}

\begin{document}
\maketitle

\begin{abstract}
The cup-to-disc ratio (CDR) is one of the most significant indicator for glaucoma diagnosis. Different from the use of costly fully supervised learning formulation with pixel-wise annotations in the literature, this study investigates the feasibility of accurate CDR measurement in fundus images using only tight bounding box supervision. For this purpose, we develop a two-task network named as CDRNet for accurate CDR measurement, one for weakly supervised image segmentation, and the other for bounding-box regression. The weakly supervised image segmentation task is implemented based on generalized multiple instance learning formulation and smooth maximum approximation, and the bounding-box regression task outputs class-specific bounding box prediction in a single scale at the original image resolution. To get accurate bounding box prediction, a class-specific bounding-box normalizer and an expected intersection-over-union are proposed. In the experiments, the proposed approach was evaluated by a testing set with 1200 images using CDR error and $F_1$ score for CDR measurement and dice coefficient for image segmentation. A grader study was conducted to compare the performance of the proposed approach with those of individual graders. The experimental results show that the proposed approach achieves CDR error of 0.0458 and $F_1$ score of 0.917 in CDR measurement and dice coefficients of 0.882 and 0.950 in optic cup and disc segmentation, respectively. These results indicate that the proposed approach outperforms the state-of-the-art performance obtained from the fully supervised image segmentation (FSIS) approach using pixel-wise annotation for CDR measurement. Its performance is also better than those of individual graders. In addition, the proposed approach gets performance close to the state-of-the-art obtained from FSIS and the performance of individual graders for optic cup and disc segmentation. The codes are available at \url{https://github.com/wangjuan313/CDRNet}.
\end{abstract}


\section{Introduction}
\label{sec:introduction}

Glaucoma is a chronic eye disease that can damage the optic nerve. It can lead to irreversible blindness if untreated. It is estimated that glaucoma will affect 111.8 million people in the world in 2040 \cite{tham2014global}. Glaucoma is called as the silence thief of sight, as its symptom is usually unaware until the very late stage. Although glaucomatous damage is irreversible, studies have shown that early diagnosis can help patients to halt or slow down its progression. The cup-to-disc ratio (CDR), comparing the vertical diameter of the optic cup (OC) portion of the optic disc (OD) with the total vertical diameter of the OD, is one of the most significant indicator for glaucoma diagnosis \cite{almazroa2015optic}. However, in clinic, it is cost-ineffective, time-consuming, subjective, and irreproducible to obtain CDR values in fundus images, since the diameters of both OC and OD are manually obtained by ophthalmologists or retinal specialists. Therefore, there is an urgent need to develop automatically computerized method for CDR measurement.

Due to the importance of CDR for glaucoma diagnosis, there have been great efforts in the literature to develop computerized methods for CDR measurement in fundus images \cite{almazroa2015optic, thakur2018survey}. Most methods, if not all, can be categorized as image-segmentation-based approach \cite{alawad2022machine}, which first segment OC and OD, and then calculate CDR values from the resulting masks. Particularly, in recent years, with the success of the deep learning in medical image analysis \cite{wang2017detecting, wang2017multi, esteva2017dermatologist, puttagunta2021medical, yu2021convolutional}, researches investigating the use of deep neural networks (DNNs) for OC and OD segmentation have dominated in CDR measurement and achieved good performance. For example, Sevastopolsky \cite{sevastopolsky2017optic} used modification of U-Net for OC and OD segmentation. Fu \textit{et al.} \cite{fu2018joint} applied multi-label DNN and polar transformation for OC and OD segmentation. Jiang \textit{et al.} \cite{jiang2019jointrcnn} employed atrous convolution, disc proposal network, cup proposal network, and the prior that OC locates inside of OD for OC and OD segmentation. Pachade \textit{et al.} \cite{pachade2021nenet} developed a nested EfficientNet  patch-based adversarial learning framework for joint OC and OD segmentation. However, all of these methods were formulated as supervised learning, which require precise pixel-wise annotations for model training. In real application, it is cost-ineffective and labor-ineffective to collect accurate pixel-wise annotations in a large scale for supervised DNN model training.

In fact, based on the definition of CDR, it is redundant to use pixel-wise annotations of OC and OD for CDR measurement. Instead, the tight bounding box annotation, defined as the smallest rectangle enclosing the whole object under consideration, is a good alternative. With tight bounding box annotation, the object must touch four sides of the bounding box and does not overlap with the region outside the bounding box, hence the height of the tight bounding box is the vertical diameter of the object. Therefore, in this study we investigate whether it is possible to use only bounding box tightness prior for accurate CDR measurement. To our best knowledge, this is the first study which applies bounding box supervision for CDR measurement in the literature.

With bounding box tightness prior, a natural solution for CDR measurement in fundus images is to develop weakly supervised image segmentation (WSIS) method using tight bounding box supervision. Some interests have been made in the literature for the development of WSIS methods using tight bounding box annotation. For example, Hsu \textit{et al.} \cite{hsu2019weakly} exploited multiple instance learning (MIL) for bounding box tightness prior and considered the well-known mask R-CNN for image segmentation. Kervadec \textit{et al.} \cite{kervadec2020bounding} leveraged the bounding box tightness prior to a deep learning via imposing a set of constraints on the network outputs for image segmentation. In our previous study \cite{wang2021bounding}, we also developed a generalized MIL formulation and smooth maximum approximation to integrate the bounding box tightness prior into DNN for image segmentation, which was extended to loose bounding box supervision using an MIL strategy based on polar transformation in \cite{wang2022polar}. While achieved good performance in image segmentation, however, compared with the fully supervised image segmentation (FSIS), these methods tend to be less accurate in the locations close to boundaries of objects, thus yielding degraded diameter values for CDR measurement. 

To resolve the problem mentioned above, in this study we develop a two-task network named as CDRNet for accurate CDR measurement, one for weakly supervised image segmentation, and the other for bounding-box regression. The weakly supervised image segmentation task is implemented based on our previous study in \cite{wang2021bounding}, in which generalized MIL formulation and smooth maximum approximation are used to incorporate the tight bounding box supervision. The bounding-box regression task outputs class-specific bounding box prediction in a single scale at the original image resolution. It is different from the traditional bounding-box regression approach used in object detectors \cite{he2017mask, lin2017focal}, which applies multiple scales and class-agnostic bounding box prediction. Such design is to accommodate the difficulty of assigning the bounding box labels to a location under consideration in the bounding-box regression optimization due to high overlap between OC and OD. To get accurate bounding box prediction, a class-specific bounding-box normalizer is introduced to obtain the bounding-box regression target; an expected intersection-over-union (eIoU) is proposed to select the positive samples for optimization, which have great potential to be highly overlapped with the bounding box labels. In the experiments, the proposed approach is evaluated by a testing set with 1200 images. A grader study is conducted to quantitatively compare the performance of the proposed approach with those of individual graders. 

In summary, the contributions of this study are as follows:
\begin{enumerate}[label=\arabic*),itemsep=0pt,topsep=0pt,parsep=0pt]
\item First, we develop CDRNet, a two-task network for accurate CDR measurement using tight bounding box supervision.
\item Second, we propose a single-scale and class-specific bounding-box regression to avoid the difficulty of assigning the bounding box labels to a location under consideration due to high overlap between OC and OD.
\item Third, we propose a bounding-box normalizer and an eIoU for bounding-box regression optimization to get accurate bounding box prediction. 
\item Finally, we conduct a grader study to compare the performance of the proposed approach with those of individual graders.
\end{enumerate}

Importantly, it has to be noted that even though this study considers CDR measurement in fundus images for glaucoma screening, the fundamental reason that the proposed approach works well for CDR measurement is due to its superior ability in accurate object size measurement (including width, height, and/or diameter). Therefore, the proposed approach is general and suitable to the applications which require accurate object size measurement. For example, in the detection of microaneurysms and dot hemorrhages in fundus images, microaneurysms and dot hemorrhages look similar in appearance, however, their diameter values are different, in which microaneurysms tend to less than 125 $\mu m$ in diameter while dot hemorrhages tend to greater than 65 $\mu m$ in diameter.

\section{Methods}

In this study, for an image under consideration, we are only provided its tight bounding box label $B$. Suppose there are $K$ bounding boxes in $B$, then it can be denoted as $B = \{\b_k, c_k\}, k = 1, 2, \cdots, K$, where the location label $\b_k=(xl_k,yt_k,xr_k,yb_k)$ is a 4-dimensional vector representing the top-left $(xl_i,yt_i)$ and bottom-right $(xr_i,yb_i)$ points of the bounding box, $c_k \in \{1,2,\cdots,C\}$ is its class label, and $C$ is the number of classes. In this study, we are interested in optic cup (OC) and optic disc (OD) classes for cup-to-disc ratio (CDR) measurement in fundus images, hence $C=2$ and $K \leq 2$.  

\subsection{CDRNet}
\subsubsection{Overview of CDRNet}

\begin{figure}[tbp] 
	\centering
	\includegraphics[trim=1in 3.05in 0.9in 2.8in,clip,width=5in]{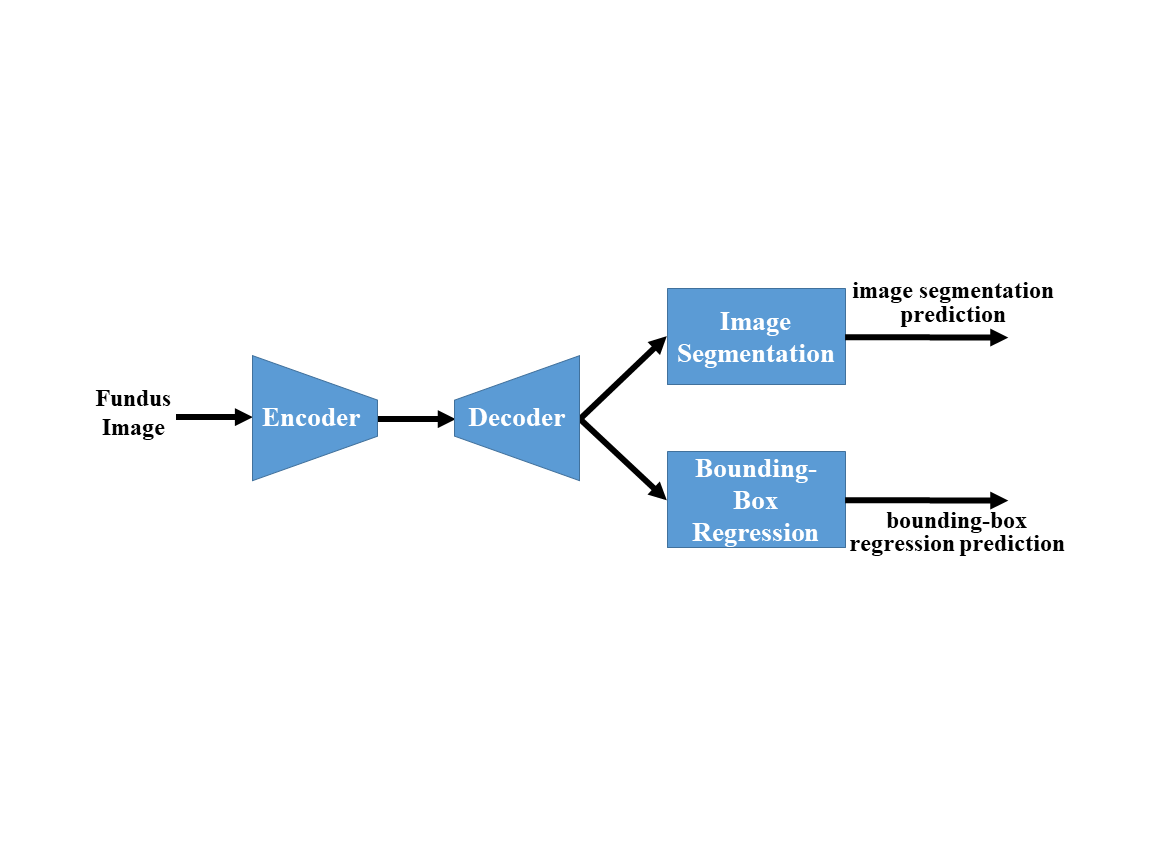}
	\caption{Illustration of the proposed network framework for accurate CDR measurement, in which encoder network, decoder network, image segmentation head, and bounding-box regression head are jointly trained in an end-to-end manner. The encoder network extracts image features at successively higher scales, the decoder network maps the image features extracted at different scales back to the original image resolution. Both image segmentation and bounding-box regression heads are feedforward neural networks.}
	\label{fig:cdrnet_overview}
\end{figure}

As noted in the introduction, compared with fully supervised image segmentation (FSIS) method, the weakly supervised image segmentation (WSIS) approach using tight bounding box supervision tend to be less accurate in the locations close to the boundaries of objects, thus yielding deteriorated CDR values. To deal with this issue, this study proposes a two-task network named as CDRNet for CDR measurement using tight bounding box supervision. The diagram of the proposed network framework is shown in Figure \ref{fig:cdrnet_overview}. It consists of a backbone (constituting an encoder network and a decoder network) and two heads (i.e. image segmentation head and bounding-box regression head), and is trained in an end-to-end manner for simultaneous OC and OD segmentation and bounding-box regression. Afterwards, a post-processing step is further applied to the predicted outputs of the two heads for CDR measurement. 

In Figure \ref{fig:cdrnet_overview}, the backbone is used to extract high-level image features for object representation, in which the encoder network is employed for extracting image features at successively higher scales, while the decoder network is applied for mapping the image features extracted at different scales back to the original image resolution \cite{wang2019hierarchical}. The image segmentation head is developed for OC and OD segmentation, taking the output of the decoder network at the original image resolution as input. In this study, the image segmentation problem is formulated as a multi-label classification problem due to the overlap between OC and OD. That is, for a location $k$ in the input image, the image segmentation head outputs a vector $\p_k$ with $C$ elements, one element for a class; each element is converted to the range of $[0,1]$ by the sigmoid function. 

More importantly, in Figure \ref{fig:cdrnet_overview}, the bounding-box regression head is proposed in this study for accurate bounding box prediction. It takes the output of the decoder network in a single scale at the original image resolution as input. Such choice is based on the small variations of diameters of both OC and OD classes in fundus images. For a location $k$ in the input image, the bounding-box regression head outputs a class-specific bounding box prediction $\v_k=[\v_{k1},\v_{k2},\cdots,\v_{kC}]$ with $4C$ elements, in which $\v_{kc}$ is a 4-dimensional vector for class $c$. This design is to accommodate the difficulty in the assignment of the bounding box labels to a location under consideration due to the high overlap between OC and OD. Note the proposed bounding-box regression in this study is different from the traditional approach used in the literature for object detection \cite{he2017mask, lin2017focal}, which takes feature maps from multiple scales as input and generates class-agnostic bounding box prediction (i.e. $\v_k$ is a 4-dimensional vector).

In the end, to obtain the optimal parameters for the proposed neural network, the two tasks are jointly trained by the multi-task loss as follows:
\begin{equation}
\mathcal{L} = \mathcal{L}_{seg} + \mathcal{L}_{reg}
\end{equation}
where $\mathcal{L}_{seg}$ is the loss for weakly supervised image segmentation with tight bounding box supervision (will be described in Section \ref{sec:method_wsis}), and $\mathcal{L}_{reg}$ is the loss for bounding-box regression (will be described in Section \ref{sec:method_regression}).

\subsubsection{Architecture of CDRNet}

\begin{figure*}[htbp] 
	\centering
	\includegraphics[trim=0in 0.4in 0in 0.4in,clip,width=5.5in]{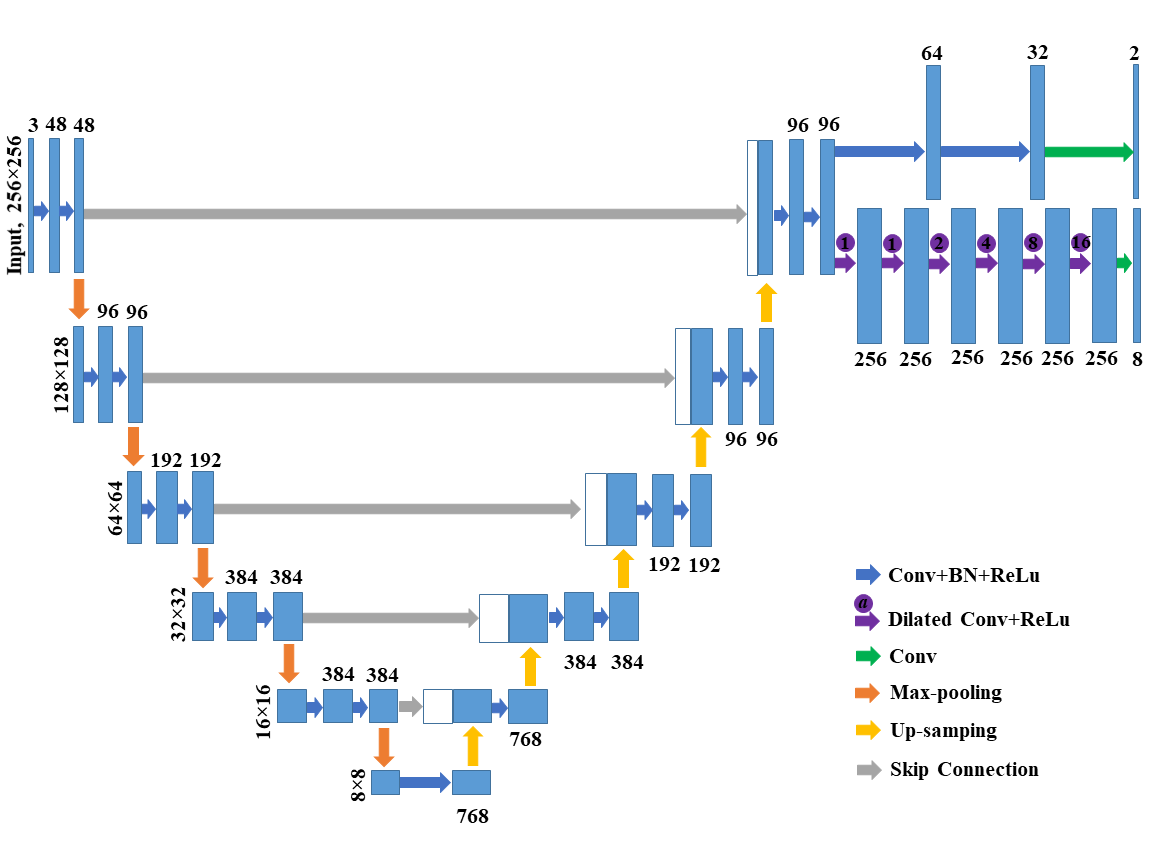}
	\caption{The architecture of CDRNet used in this study for simultaneous OC and OD segmentation and CDR measurement. The dimension of the input image is set to be $256\times256$ in this plot as example. ``\textbf{Conv + BN + ReLu}`` denotes the cascade of Conv, BN, and ReLU layers. ``\textbf{Dilated Conv + ReLu}`` indicates the cascade of dilated Conv and ReLU layers, in which the dilation factor is denoted by $a$. For each feature map, the number of its channel is shown at the top or bottom of its location. }
	\label{fig:cdrnet_architecture}
\end{figure*}

Based on the proposed framework in Figure \ref{fig:cdrnet_overview}, the architecture of CDRNet used in this study is shown in detail in Figure \ref{fig:cdrnet_architecture}. The structures of the encoder and decode networks are designed based on U-net \cite{ronneberger2015u}. The encoder network is formed by a cascade of convolutional (Conv) layers, batch normalization (BN) layers, rectified linear unit (ReLU) layers, and max-pooling layers. Similarly, the decoder network is formed by a cascade of Convs, BNs, ReLUs, up-sampling layers, and skip-connection units. The skip connections are used to combine information from deeper layers with that from a shallow layer. The skip connections are important components adopted in U-net and have been shown to be able to improve image segmentation performance. The image segmentation head is a traditional feed-forward neural network, which consists of a cascade of Convs, BNs, and ReLUs. The bounding-box regression head is developed based on dilated convolutional network \cite{yu2015multi} to accommodate the large size of OC and OD. It constitutes a cascade of dilated Convs and ReLUs with gradually increasing dilation factors. The dilated Convs are used in this study to ensure large receptive field for bounding-box regression while keeping the output dimension of the network to be same as the input dimension of the network.

In Figure \ref{fig:cdrnet_architecture}, the kernel size is set to be $3\times3$ for all of the Convs. All of the max-pooling layers are set to have kernel size $2\times2$ with stride 2. Correspondingly, the scale factor is set to 2 for all of the up-sampling layers. The dilation factors of the six dilated Convs are set as 1, 1, 2, 4, 8, and 16, respectively. Importantly, it has to be noted that in Figure \ref{fig:cdrnet_architecture}, the input image size has to be divisible to 32 due to the use of five max-pooling layers. 

\subsection{Weakly supervised image segmentation}
\label{sec:method_wsis}

In CDRNet, the weakly supervised image segmentation (WSIS) approach developed in our previous study \cite{wang2021bounding} is considered for OC and OD segmentation. It utilizes generalized multiple instance learning (MIL) and smooth maximum approximation to integrate the bounding box tightness prior. We simply describe the WSIS approach here and recommend reference \cite{wang2021bounding} for more details. 

\subsubsection{MIL formulation}
\label{sec:method_wsis_mil}

\begin{figure}[htbp] 
	\centering
	\includegraphics[trim=0in 0in 0in 0in,clip,width=3in]{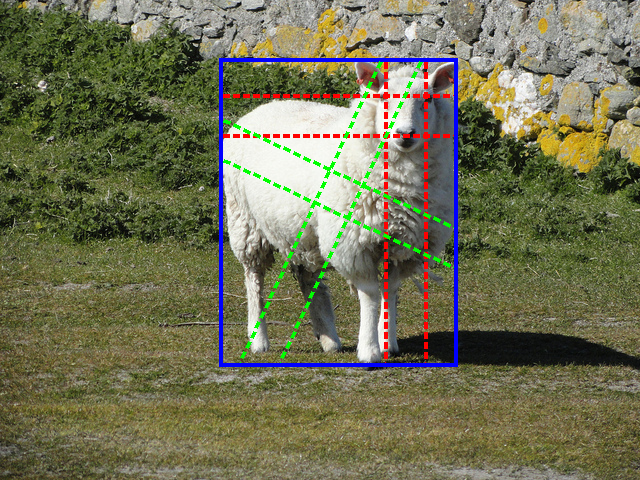}
	\caption{Demonstration of positive bags, in which the blue rectangle is the tight bounding box of the object, the green dashed lines are examples of positive bags with $\theta=25^\circ$, and the red dashed lines are examples of positive bags with $\theta=0^\circ$.}
	\label{fig:mil_demonstration}
\end{figure}

The crossing line of a bounding box is defined as a line with its two endpoints located on the opposite sides of the box. Considering an object in class $c$, any crossing line in the tight bounding box has at least one pixel belonging to the object in the box, thus pixels on a cross line compose a positive bag for class $c$. Based on this observation, in this study the positive bags are defined as all parallel crossing lines obtained at a set of different angles. An parallel crossing line is parameterized by an angle $\theta \in (-90^\circ, 90^\circ)$ with respect to the edges of the box where its two endpoints locate. For an angle $\theta$, two sets of parallel crossing lines are obtained, one touches top and bottom edges of the box, and the other crosses left and right edges of the box. For example, in Figure~\ref{fig:mil_demonstration}, we show examples of positive bags obtained at two different angles, in which those marked by green and red dashed lines have $\theta=25^\circ$ and $\theta=0^\circ$, respectively.

Moreover, in an image under consideration, pixels outside of any bounding boxes in class $c$ do not belong to class $c$, thus pixels outside of any bounding boxes in class $c$ are negative bags. In this study, for class $c$, we consider all of individual pixels outside of any bounding boxes in class $c$ as negative bags, in which each individual pixel is defined as a negative bag. 

\subsubsection{Weakly supervised segmentation loss}
\label{sec:method_wsis_loss}

To optimize the segmentation task, this study considers MIL loss with two terms. For class $c$, suppose its positive and negative bags are denoted by $\mathcal{B}_c^+$ and $\mathcal{B}_c^-$, respectively, then MIL loss $\mathcal{L}_c$ is:
\begin{equation}
\mathcal{L}_c = \phi_c(P; \mathcal{B}_c^+, \mathcal{B}_c^-) + \lambda \varphi_c(P)
\label{equ:mil_loss}
\end{equation}
where $\phi_c$ is the unary loss, $\varphi_c$ is the pairwise loss, $P$ is the image segmentation prediction, and $\lambda$ is a constant controlling the trade off between the two losses.

The unary loss $\phi_c$ enforces bounding box tightness constraint on the image segmentation prediction $P$ for both positive bags $\mathcal{B}_c^+$ and negative bags $\mathcal{B}_c^-$. In MIL formulation, a positive bag of class $c$ contains at least one pixel inside the object of class $c$, thus the pixel with maximum prediction in the positive bag tends to be positive sample for class $c$; no pixels in a negative bag belong to any objects of class $c$, thus even the pixel with maximum prediction in the negative bag is negative sample for class $c$. Based on these observations, the unary loss $\phi_c$ is defined as a focal loss \cite{lin2017focal} for bag prediction as follows:
\begin{equation}
\begin{split}
\phi_c = -\frac{1}{N^+} \left( \sum_{b \in \mathcal{B}_c^+} \beta \left(1-P_c(b)\right)^\gamma \log P_c(b) + \right. 
\\ 
\left. \sum_{b \in \mathcal{B}_c^-} (1-\beta)P_c(b)^\gamma \log(1-P_c(b)) \right)
\label{equ:focal_loss}
\end{split}
\end{equation}
where $P_c(b) = \max_{k \in b}(p_{kc})$ is the image segmentation prediction of the bag $b$ being positive for class $c$, $p_{kc}$ is the image segmentation prediction of the \textit{k}th location for class $c$, $N^+ = \max(1, \abs{\mathcal{B}_c^+})$, $\abs{\mathcal{B}}$ is the cardinality of $\mathcal{B}$, $\beta \in [0,1]$ is the weighting factor, and $\gamma \geq 0$ is the focusing parameter. The unary loss gets minimum when $P_c(b)=1$ for positive bags and $P_c(b)=0$ for negative bags. 

The pairwise loss $\varphi_c$ is designed to pose the piece-wise smoothness on the image segmentation prediction. It is defined as:
\begin{equation}
\varphi_c = \frac{1}{\abs{\varepsilon}} \sum_{(k,k^\prime) \in \varepsilon} \left( p_{kc} - p_{k^\prime c} \right) ^2 
\end{equation}
where $\varepsilon$ is the set containing all neighboring pixel pairs. 

Finally, considering all of the $C$ classes, the loss $\mathcal{L}_{seg}$ for weakly supervised image segmentation is:
\begin{equation}
\mathcal{L}_{seg} = \sum_{c=1}^C \mathcal{L}_c
\label{equ:seg_loss}
\end{equation}

\subsubsection{Smooth maximum approximation}
\label{sec:method_wsis_smooth_max}

In the unary loss, the maximum prediction of pixels in a bag is used as bag prediction $P_c(b)$. However, the derivative $\partial P_c / \partial p_{kc}$ is discontinuous, resulting in numerical instability. Moreover, the derivative $\partial P_c / \partial p_{kc}$ has value 0 for all but the maximum $p_{kc}$, thus optimizes only one pixel in the bag. This is suboptimal considering that there are usually more than one pixel belonging to the object in most positive bags. To alleviate these issues, we propose to replace the maximum function by its smooth maximum approximation \cite{lange2014applications}. Specifically, for the maximum function $f(\x) = \max_{i=1}^n x_i$, two variants of its smooth maximum approximation as follows are considered. 

(1) \textit{$\alpha$-softmax function:}
\begin{equation}
S_{\alpha}(\x) = \frac{\sum_{i=1}^n x_i e^{\alpha x_i}}{\sum_{i=1}^n e^{\alpha x_i}}
\end{equation}
where $\alpha>0$ is a constant value. The higher $\alpha$ value yields better approximation. 

(2) \textit{$\alpha$-quasimax function:}
\begin{equation}
Q_{\alpha}(\x) = \frac{1}{\alpha} \log \left(\sum_{i=1}^n e^{\alpha x_i}\right) - \frac{\log n}{\alpha}
\end{equation}
where $\alpha>0$ is a constant value. The higher $\alpha$ value also gets better approximation. 

\subsection{Bounding-box regression}
\label{sec:method_regression}

\subsubsection{Bounding-box regression target}
\label{sec:method_regression_target}

For bounding-box regression, only the locations which fall into at least one bounding box label are considered as positive samples for optimization, and those outside of any bounding box labels are ignored during optimization. For a location of positive sample, the bounding-box regression head predicts the normalized shift to this location for each class. In bounding-box regression optimization, for each class, at most one matched bounding box label is first assigned to this location, and then the assigned matched bounding box label is used to obtain its bounding-box regression target. 

For a location under consideration, its matched bounding box label at class $c$ is determined as follows: 1) if the location is within a bounding box label of class $c$, then this bounding box label is assigned as its matched bounding box label; 2) if the location is in multiple bounding box labels of class $c$, then the $L_1$ norm of the bounding-box regression target with respect to each bounding box label is calculated, and the one with minimum $L_1$ norm is assigned as the matched bounding box label. Let the location be $(x_i,y_i)$, suppose its matched bounding box label at class $c$ is $\b=(xl,yt,xr,yb)$ of class $c$, then its bounding-box regression target $\t_{ic}=(tl_{ic}, tt_{ic}, tr_{ic}, tb_{ic})$ is defined as:
\begin{equation}
\renewcommand\arraystretch{1.3}
\begin{array}{cc}
tl_{ic} = (x_i - xl)/S_{c} & tt_{ic} = (y_i - yt)/S_{c} \\
tr_{ic} = (xr - x_i)/S_{c} & tb_{ic} = (yb - y_i)/S_{c} \\
\end{array}
\label{equ:regression_targets}
\end{equation}
where $S_c$ is the normalizer of the object size for class $c$. In this study, it is estimated as the average of the mean vertical diameter and mean horizontal diameter of objects in class $c$. 

\subsubsection{Positive sample selection}
\label{sec:method_sample_selection}

Among the positive samples obtained above, those far from the centers of the matched bounding box labels yield large bounding-box regression target, which might adversarially affect the bounding-box regression optimization. Besides, the above positive sample definition yields $WH$ positive samples for a non-overlapped matched bounding box label with width $W$ and height $H$. Such large number of correlated positive samples are redundant for optimization. To solve these issues, this study proposes an expected intersection-over-union (eIoU) for positive sample selection. For a location of positive sample, its eIoU is defined as the maximum IoU between its matched bounding box label and all of the possible predicted bounding boxes centered at this location. Upon the definition of eIoU, the positive samples satisfying $eIoU{>}T$ are selected for bounding-box regression optimization and those with $eIoU{\leq}T$ are ignored during optimization, where the threshold $T$ is a constant in the range of $[0, 1]$.

\begin{figure}[htbp] 
	\centering
	\includegraphics[trim=2.4in 2in 3in 1.8in,clip,width=3.5in]{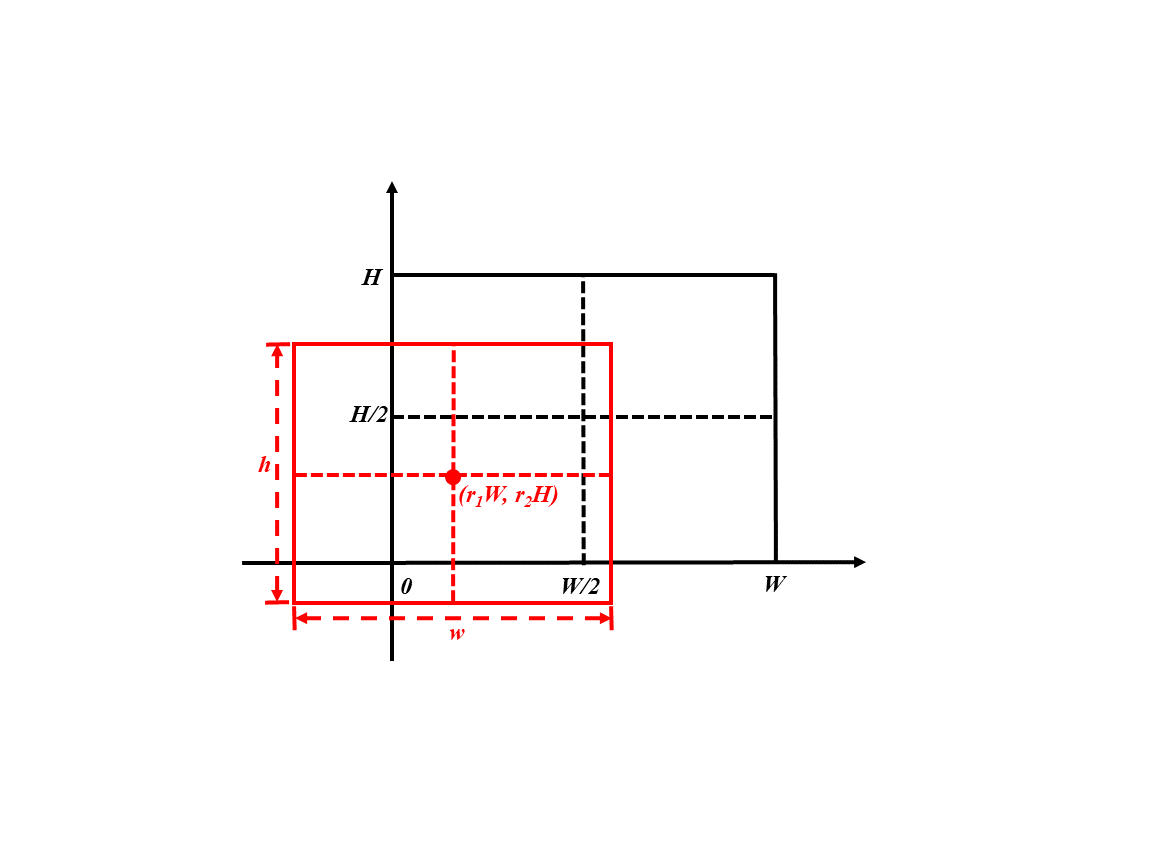}
	\caption{Illustration of eIoU calculation. In this plot, the red dot gives the location of positive sample, the black rectangle is its matched bounding box label, and the red rectangle in dashed line is an example of bounding box centered at the location of positive sample.}
	\label{fig:expected_iou}
\end{figure}

In Figure \ref{fig:expected_iou}, we give an illustration of eIoU calculation. In this plot, the location of positive sample is denoted by the red dot. Its matched bounding box label is marked by the black rectangle of width $W$ and height $H$. For simplicity, the location of positive sample is expressed as $(r_1W,r_2H)$, where $0 < r_1,r_2 < 1$. Centered at this location, it is able to construct a set of bounding boxes with different width $w$ and height $h$. One example is shown as the red rectangle in Figure \ref{fig:expected_iou}. By varying $w$ and $h$, we are able to get a set of IoUs between the black rectangle and the red rectangle, and the maximum IoU is the eIoU of this location. 

For $0 < r_1,r_2 \leq 0.5$, it can be easily proved that the eIoU is obtained at one of the following four conditions: 
\begin{enumerate}[label=\arabic*),itemsep=1pt]
\item $w=2r_1W$, $h=2r_2H$
\item $w=2r_1W$, $h=2(1-r_2)H$
\item $w=2(1-r_1)W$, $h=2r_2H$
\item $w=2(1-r_1)W$, $h=2(1-r_2)H$
\end{enumerate}
The corresponding IoUs are:
\begin{equation}
\renewcommand\arraystretch{1.5}
\begin{array}{ll}
IoU_1(r_1,r_2) = 4r_1r_2, & \mbox{for cond. 1)}  \\
IoU_2(r_1,r_2) = 2r_1/(2r_1(1-2r_2)+1), & \mbox{for cond. 2)} \\
IoU_3(r_1,r_2) = 2r_2/(2r_2(1-2r_1)+1), & \mbox{for cond. 3)} \\
IoU_4(r_1,r_2) = 1/(4(1-r_1)(1-r_2)), & \mbox{for cond. 4)} \\
\end{array}
\label{equ:eiou_condition}
\end{equation}
In the end, the eIoU for $0 < r_1,r_2 \leq 0.5$ is as follows:
\begin{equation}
eIoU(r_1,r_2) =  \max_{i=1,2,3,4} IoU_i\left(r_1, r_2\right)
\label{equ:eiou_1stquandrant}
\end{equation}

Moreover, one can prove that for $0.5 \leq r_1 < 1$, eIoU is obtained by replacing $r_1$ in equation \eqref{equ:eiou_1stquandrant} with $1-r_1$; for $0.5 \leq r_2 <1$, eIoU is obtained by replacing $r_2$ in equation \eqref{equ:eiou_1stquandrant} with $1-r_2$. Therefore, the eIoU can be obtained as follows:
\begin{equation}
\begin{split}
eIoU(r_1,r_2) =  \max_{i=1,2,3,4} IoU_i\left(\min(r_1,1-r_1), \right. \\
\left. \min(r_2,1-r_2)\right)
\end{split}
\label{equ:eiou}
\end{equation}

\subsubsection{Bounding-box regression loss}
\label{sec:method_regression_loss}

To optimize the bounding-box regression task, the smooth $L_1$ loss is considered. For a location $(x_i,y_i)$, let its bounding-box regression prediction for class $c$ be $\v_{ic}$ and its bounding-box regression target be $\t_{ic}$, the loss for bounding-box regression is defined as:
\begin{equation}
\mathcal{L}_{reg} = \frac{1}{C} \sum_{c=1}^C \frac{1}{M_c} \sum_{i=1}^{M_c} s(\t_{ic}-\v_{ic})
\label{equ:reg_loss}
\end{equation}
where $M_c$ is the number of positive samples for class $c$, $s(\x)$ is the summation of the smooth $L_1$ loss for all elements of $\x$. For an element $x$, its smooth $L_1$ loss is 
\begin{equation}
SL_1(x) =
\begin{cases}
\frac{1}{2}(\sigma x)^2 & \text{if $|x|<\frac{1}{\sigma^2}$} \\
|x| - \frac{1}{2\sigma^2} & \text{otherwise} \\
\end{cases}
\label{equ:smoothL1_func}
\end{equation}
where $\sigma$ is a parameter controlling the switch from $L_2$ loss to $L_1$ loss. 

\subsection{Post-processing for CDR measurement}

With both image segmentation prediction and bounding-box regression prediction, the CDR value is calculated as follows: first, for each class, the bounding-box regression prediction corresponding to highest value in image segmentation prediction is selected. For simplicity, let the bounding-box regression prediction for OC and OD be $\v_{oc}$ and $\v_{od}$, respectively. The predicted location of OC $\hat{\b}_{oc}=(\hat{xl}_{oc},\hat{yt}_{oc},\hat{xr}_{oc},\hat{yb}_{oc})$ is calculated using the inverse of equation \eqref{equ:regression_targets} by replacing $\t_{ic}$ with $\v_{oc}$. Similarly, the predicted location of OD $\hat{\b}_{od}=(\hat{xl}_{od},\hat{yt}_{od},\hat{xr}_{od},\hat{yb}_{od})$ is obtained from $\v_{od}$. Finally, CDR is calculated as follows:
\begin{equation}
C\!D\!R = (\hat{yb}_{oc}-\hat{yt}_{oc})/(\hat{yb}_{od}-\hat{yt}_{od})
\label{equ:cdr_calculation}
\end{equation}

\section{Experiments}

\subsection{Dataset}
\label{sec:experiment_dataset}

\begin{figure}[htbp] 
	\centering
	\setlength{\tabcolsep}{2pt}
	\begin{tabular}{cccc}
	\includegraphics[trim=0in 0in 0in 0in,clip,width=1in]{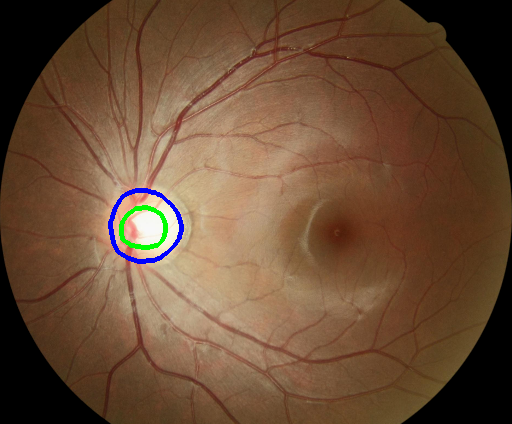} &
	\includegraphics[trim=0in 0in 0in 0in,clip,width=1in]{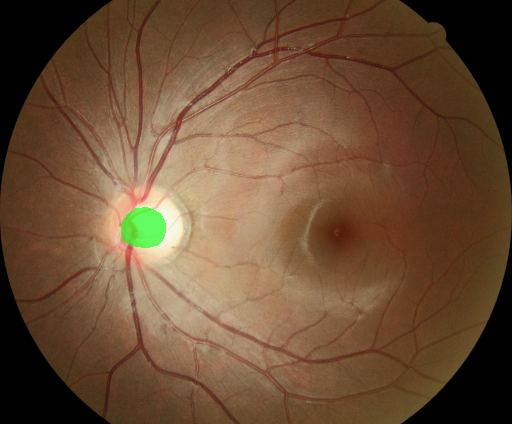} &
	\includegraphics[trim=0in 0in 0in 0in,clip,width=1in]{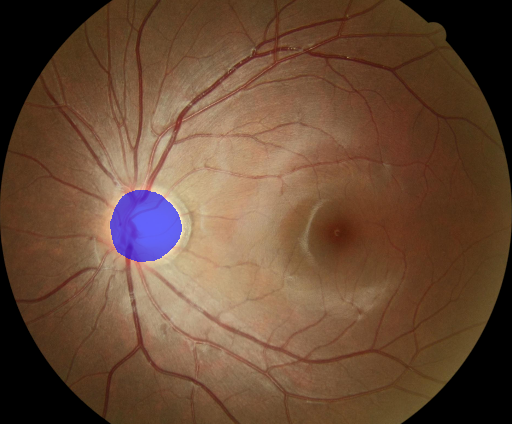} &
	\includegraphics[trim=0in 0in 0in 0in,clip,width=1in]{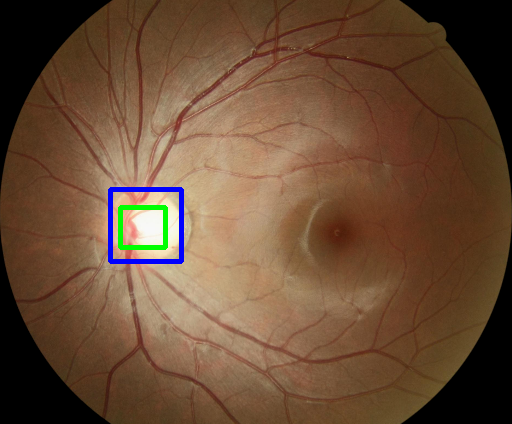} \\
	\includegraphics[trim=0in 0in 0in 0in,clip,width=1in]{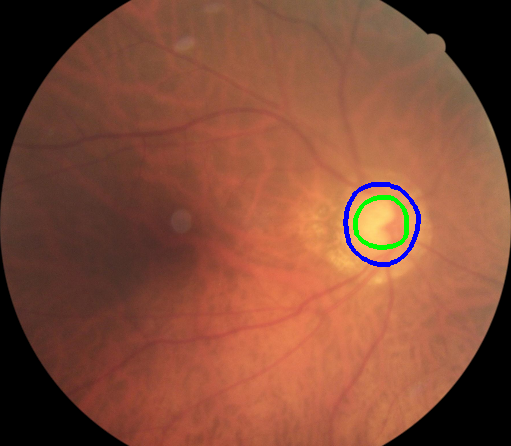} &
	\includegraphics[trim=0in 0in 0in 0in,clip,width=1in]{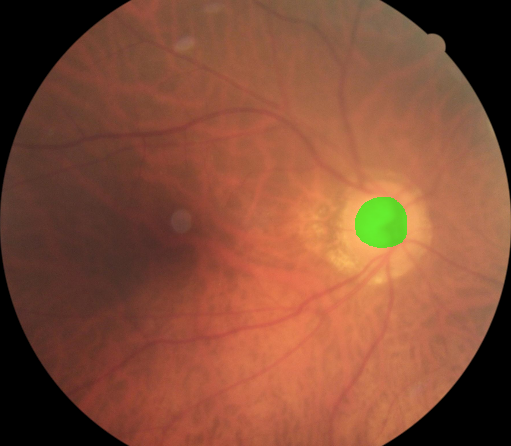} &
	\includegraphics[trim=0in 0in 0in 0in,clip,width=1in]{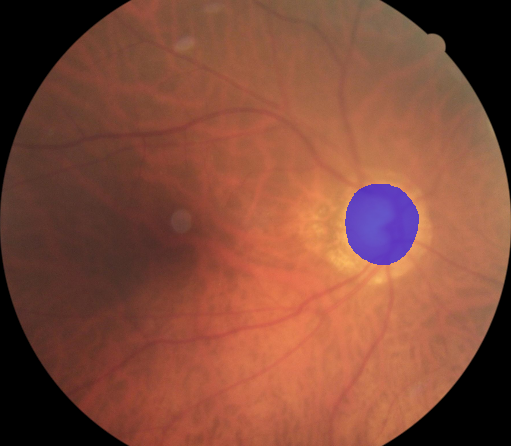} &
	\includegraphics[trim=0in 0in 0in 0in,clip,width=1in]{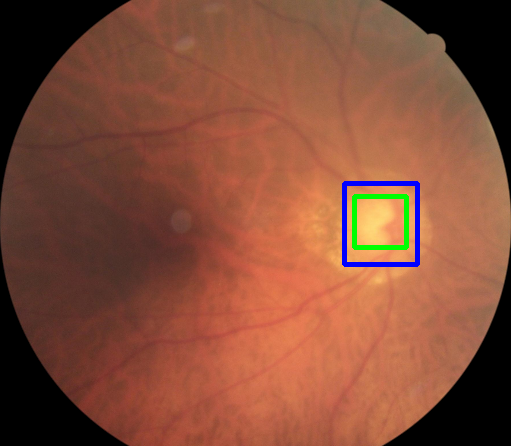}\\
	\includegraphics[trim=0in 0in 0in 0in,clip,width=1in]{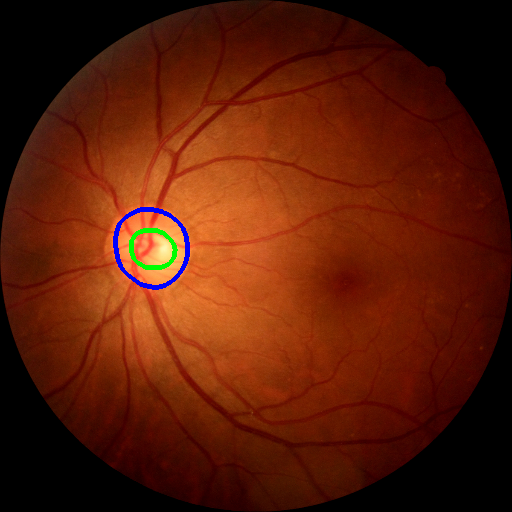} &
	\includegraphics[trim=0in 0in 0in 0in,clip,width=1in]{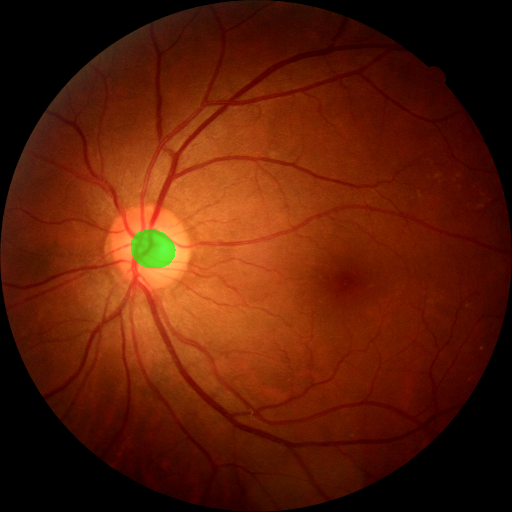} &
	\includegraphics[trim=0in 0in 0in 0in,clip,width=1in]{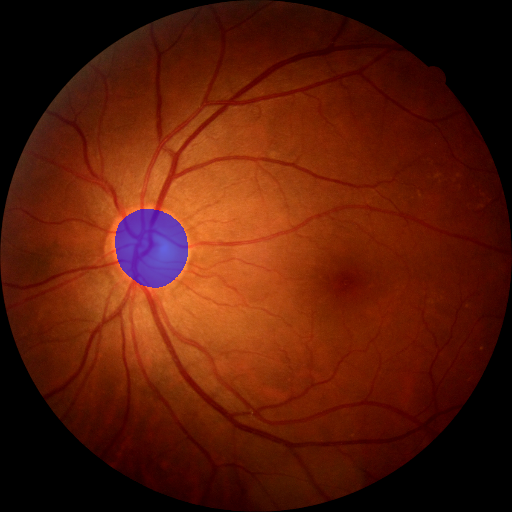} &
	\includegraphics[trim=0in 0in 0in 0in,clip,width=1in]{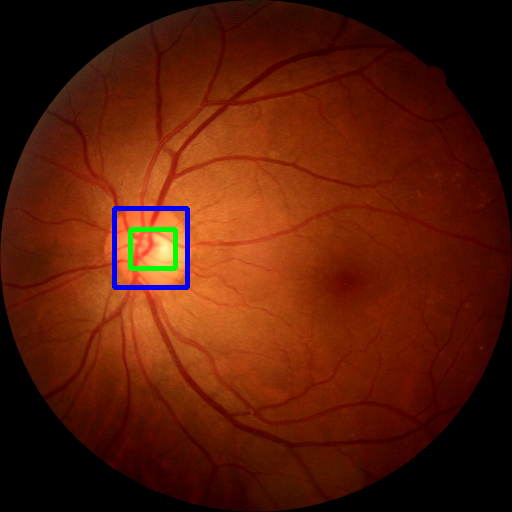}\\ 
	\includegraphics[trim=0in 0in 0in 0in,clip,width=1in]{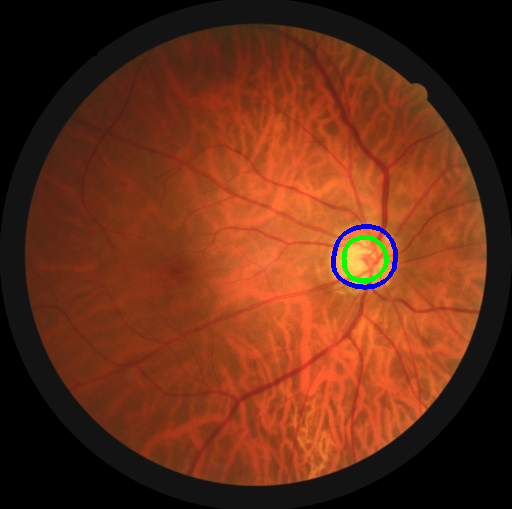} &
	\includegraphics[trim=0in 0in 0in 0in,clip,width=1in]{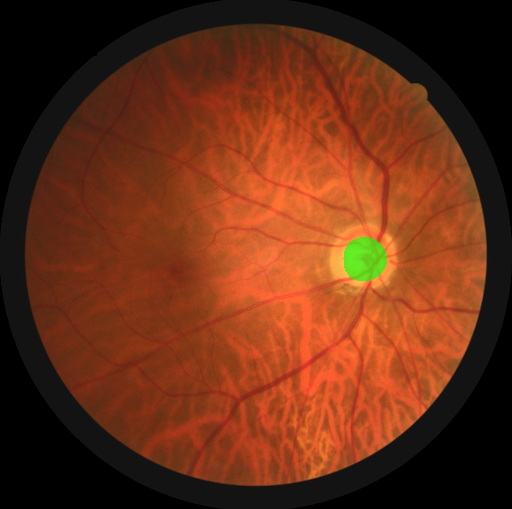} &
	\includegraphics[trim=0in 0in 0in 0in,clip,width=1in]{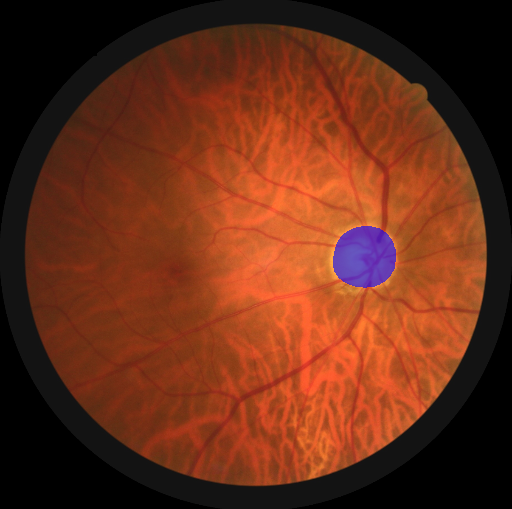} &
	\includegraphics[trim=0in 0in 0in 0in,clip,width=1in]{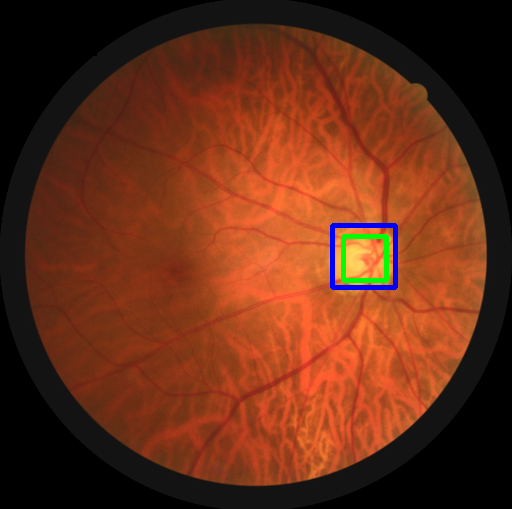}\\ 
	(a) & (b) & (c) & (d) \\
	\end{tabular}
	\caption{Examples of fundus images and their annotations, including (a) boundaries of OC and OD provided by retinal technicians, (b) mask of OC for supervised learning, (c) mask of OD for supervised learning, (d) tight bounding box labels of OC and OD for weakly supervised learning. In these plots, OC and OD are indicated by green and blue colors, respectively.}
	\label{fig:dataset_examples}
\end{figure}

This study made use of a total of 8881 digital fundus images in the experiments, all of which were collected by Shenzhen SiBright Co., Ltd. (Shenzhen, Guangdong, China). They were randomly divided into three non-overlapped subsets, one with 7081 images for training, one with 600 images for validation, and one with 1200 images for testing. All images in the dataset were graded once by a group of four retinal technicians (denoted as $G_1$, $G_2$, $G_3$, and $G_4$), who were trained at least six months for OC and OD identification before grading. During grading, the retinal technicians were required to provide the boundaries of both OC and OD, the corresponding masks of which were used as ground truth for fully supervised training in the experiments. The tight bounding box labels were converted from the boundaries of OC and OD, which were used as ground truth for weakly supervised training in the experiments. Taking four fundus images as examples, we show boundaries of OC and OD provided by retinal technicians in Figure \ref{fig:dataset_examples}(a), their corresponding masks of OC and OD for supervised learning in Figures \ref{fig:dataset_examples}(b) and \ref{fig:dataset_examples}(c), and their corresponding tight bounding box labels of OC and OD for weakly supervised learning in Figure \ref{fig:dataset_examples}(d).

\subsection{Grader study}
To compare the performance of the proposed CDRNet with those of individual graders, a grader study was conducted on an independent dataset with 229 digital fundus images. The same four retinal technicians $G_1$, $G_2$, $G_3$, and $G_4$ participated in the grader study. They independently grade the boundaries of both OC and OD in the fundus images, yielding four readings for each image in this dataset. Therefore, four sets of results are obtained for each image in this dataset, each set consists of an OC mask, an OD mask, and a CDR value derived from the reading of a grader. For example, for an image under consideration, its first set of result is obtained from the boundaries of OC and OD provided by $G_1$ as follows: OC and OD masks are obtained as the close region of the boundaries of OC and OD, separately; the CDR value is calculated from the height of the corresponding tight bounding box labels of OC and OD.

In the grader study analysis, the proposed CDRNet model is considered as a virtual grader, yielding five graders in total. For a performance metric under consideration, the performance values of all graders are provided in a pairwise fashion, resulting in a $5\times5$ performance table. In this table, four performance values for each grader are reported when the readings from the other graders are treated as ground truth \cite{wang2019feasibility}. To summarize the performance of an individual grader, the average of its four performance values is used \cite{wang2019feasibility}. 

\subsection{Performance evaluation}
\subsubsection{Performance evaluation for CDR measurement}
To evaluate the performance of the proposed CDRNet for CDR measurement, we consider the CDR error, which is defined as the absolute difference between the predicted CDR value obtained from CDRNet and that from the corresponding bounding box label. Mathematically, it can be formatted as:
\begin{equation}
\mbox{CDR error} = \lvert (yb_{oc}-yt_{oc})/(yb_{od}-yt_{od}) - (\hat{yb}_{oc}-\hat{yt}_{oc})/(\hat{yb}_{od}-\hat{yt}_{od}) \lvert
\end{equation}
where $yb_{oc}$ and $yt_{oc}$ are the top and bottom locations of the bounding box label of OC, $yb_{od}$ and $yt_{od}$ are the top and bottom locations of the bounding box label of OD, and $\hat{y}$ is the prediction of $y$ obtained from network. The smaller CDR error indicates better performance in CDR measurement. CDR error is zero when the predicted CDR value is perfect. 

In glaucoma screening, $C\!D\!R \geq 0.6$ is usually used as diagnosis standards to screen the suspect glaucoma \cite{gupta2016prevalence}, hence the images with $C\!D\!R \geq 0.6$ are diagnosed as positive and those with $C\!D\!R < 0.6$ are negative. To measure the performance of the proposed network for glaucoma screening, $F_1$ score is considered. It is a performance summary which conveys the balance between precision (or positive predictive value, PPV) and recall (or true positive rate, TPR), the most important metrics for glaucoma screening. Mathematically, $F_1$ score is defined as the harmonic mean of precision and recall as follows:
\begin{equation}
F_1 = 2 \times \frac{PPV \times TPR}{PPV + TPR} = \frac{2TP}{2TP+FP+FN}
\end{equation}
where $TP$, $FP$ and $FN$ are true positive, false positive, and false negative for suspect glaucoma diagnosis, separately.
$F_1$ score is in the range of [0, 1]. The higher $F_1$ score suggests better performance. 

\subsubsection{Performance evaluation for OC and OD segmentation}
To measure the performance of the proposed CDRNet for OC and OD segmentation, the dice coefficient was employed, which has been widely used as performance metric for image segmentation. The dice coefficient is defined as:
\begin{equation}
dice = 2TP/(2TP + FP + FN)
\end{equation}
where $TP$, $FP$ and $FN$ are true positive, false positive, and false negative for image segmentation, separately.
The dice coefficient is in the range of $[0, 1]$. The higher dice value indicates better segmentation performance.

\subsection{Methods for comparison}
In the experiments, we evaluated the performance of the proposed CDRNet for CDR measurement and OC and OD segmentation in fundus images. We also compared it against the FSIS approach \cite{ronneberger2015u}, which has been widely used for CDR measurement in fundus images and achieves the state-of-the-art performance for both CDR measurement and OC and OD segmentation. For the proposed approach using weakly supervised learning based on tight bounding box annotations, the FSIS approach can be served as the upper bound of its performance for both CDR measurement and OC and OD segmentation due to the use of fully supervised learning based on costly pixel-wise annotations. Furthermore, we considered two methods for comparison, both of which also consider tight bounding box supervision. The first method is the WSIS approach developed in \cite{wang2021bounding}. It has been demonstrated to outperform several WSIS methods in the literature and achieves segmentation performance close to the FSIS approach. This WSIS approach is considered in this study to demonstrate the benefits of introducing bounding-box regression in the proposed network. The second method is RetinaNet \cite{lin2017focal}. It is a well-known single-stage object detector, which has got the state-of-the-art performance in object detection. RetinaNet conducts only box regression for object detection, thus cannot make the image segmentation prediction. It is considered in this study to investigate the advantages of including weakly supervised image segmentation in the proposed approach.

In FSIS and WSIS, the predicted CDR values was calculated from the predicted OC and OD masks, by first converting the predicted masks into bounding boxes. Similar as the proposed network, RetinaNet adopts the bounding-box regression predictions with highest classification scores to get CDR values. Note RetinaNet is an object detector, thus there are no image segmentation results.

For fairness of comparison, in the experiments, the network structure used in FSIS and WSIS is same, which was obtained by removing the bounding-box regression head from the architecture of the proposed CDRNet in Figure \ref{fig:cdrnet_architecture}. For RetinaNet, the backbone is set as the encoder network in Figure \ref{fig:cdrnet_architecture} and the other settings are same as those in \cite{lin2017focal}.

\subsection{Implementation details}
In this study, all experiments were implemented using PyTorch in four GPU cards of GeForce GTX TITAN X with 12 GB memory. As a preprocessing step, all of the fundus images were resized to have width of 512 pixels in the experiments. To save the memory in model training, an online image cropping operation was applied to the images in the training set to obtain region of interests (ROIs) with dimension of $256\times256$ pixels as network input. For effective training, the image cropping operation was set such that at least 75\% of ROIs contain OC and/or OD regions. No image cropping was employed for images in the validation and testing sets. 

For model training, the Adam optimizer \cite{kingma2014adam} was used with following parameter values: initial rate 0.001, $\beta_1=0.9$, and $\beta_2 = 0.999$. The learning rate decreases by a factor of 10 if the validation loss does not decrease in 4 epochs. The batch size was set to consume as much GPU memory as possible, which was 32 for the proposed approach, 96 for FSIS and WSIS, and 160 for RetinaNet. To speedup the training process, an early stop criterion was applied, which terminates the training if the validation loss does not decrease in 8 epochs.

The bounding-box normalizer in equation \eqref{equ:regression_targets} was set as $S_{oc}=40$ for OC class and $S_{od}=70$ for OD class, which were estimated by all of the objects in the training set. The parameter in equation \eqref{equ:mil_loss} was set to be $\lambda=10$ based on our previous study \cite{wang2021bounding}. The parameters in unary loss \eqref{equ:focal_loss} were set as $\beta=0.25$ and $\gamma=2$ according to the focal loss in study \cite{lin2017focal}. The parameter $\theta$ in positive bag definition, $\alpha$ in smooth maximum approximation, $\sigma$ in smooth $L_1$ loss, and $T$ in positive sample selection were all selected by grid search. In particular, the following parameter values were considered in the experiments for grid search: $\theta \in \{(-40^\circ,40^\circ,10^\circ), (-40^\circ,40^\circ,20^\circ),(-60^\circ,60^\circ,30^\circ)\})$, $\alpha \in \{6,8,10\}$, $\sigma \in \{3,4,5,6,7,8\}$, and $T \in \{0.5, 0.6, 0.7\}$. The optimal values are $\theta=(-40^\circ,40^\circ,10^\circ)$, denoting evenly spaced angle values within interval $(-40^\circ,40^\circ)$ with step $10^\circ$, $\alpha=8$, $\sigma=6$, $T=0.6$ for $\alpha$-softmax function, and $T=0.5$ for $\alpha$-quasimax function. The parameters $\theta$ and $\alpha$ were selected based on the WSIS model, and then simply used by the proposed CDRNet model without further optimization. 

To further enlarge the set of images for training, an on-line data augmentation procedure \cite{wang2018context, wang2020simultaneous} as follows was applied to the fundus images in the training set: (1) randomly flip images from left to right with probability of 0.5, (2) randomly flip images up down with probability of 0.5, (3) zoom in/out images with scaling factor randomly selected in the range of [0.95, 1.05], (4) randomly translate images along both dimensions by at most 5\% pixels along that dimension, (5) adjust the image contrast with random enhancement factor in the range of [0.9, 1.1], (6) adjust the image brightness with random factor in the range of [-0.1, 0.1], (7) adjust the hue of images with random factor in the range of [-0.05, 0.05], and (8) adjust the saturation of images with random enhancement factor in the range of [0.95, 1.05]. 

\section{Results}
\label{sec:results}

\subsection{Performance comparison of the proposed CDRNet and different existing methods}

\subsubsection{Main results: performance of CDR measurement}

In Table \ref{table:main}, we show the results of the proposed CDRNet on the testing set for CDR measurement, in which the performance is measured by CDR errors and $F_1$ scores. Two models of the proposed approach are considered, one adopting $\alpha$-softmax function and the other using $\alpha$-quasimax function. For comparison, the results obtained from the FSIS model are also shown in Table \ref{table:main}. As can be seen, the CDR errors and $F_1$ scores of both models of the proposed CDRNet are close to or even better than those of the FSIS model, indicating similar or better performance of the proposed approach in CDR measurement. In particular, the proposed approach using $\alpha$-softmax approximation function achieves CDR error of 0.0458 and $F_1$ score of 0.917, compared to CDR error of 0.0465 and $F_1$ score of 0.911 for the FSIS model. 

\begin{table}
\caption{Performance comparison of the proposed CDRNet and different existing methods in CDR measurement. The methods as follows are considered for comparison: 1) FSIS, 2) WSIS, and 3) RetinaNet.}
\centering
\begin{tabular}{ccc}
\hline
\hline
 & CDR error & $F_1$ score \\
\hline
CDRNet ($\alpha$-softmax) & \textbf{0.0458} & \textbf{0.917} \\
CDRNet ($\alpha$-quasimax) & 0.0468 & 0.912 \\
FSIS & 0.0465 & 0.911 \\
WSIS ($\alpha$-softmax) & 0.0623 & 0.890 \\
WSIS ($\alpha$-quasimax) & 0.0600 & 0.899 \\
RetinaNet & 0.0468 & 0.903 \\
\hline
\hline
\end{tabular}
\label{table:main}
\end{table}

To demonstrate the benefits of introducing bounding-box regression in the proposed approach, CDR errors and $F_1$ scores obtained by the WSIS models are also given in Table \ref{table:main} for comparison. Similarly, two WSIS models are considered in the experiments, one using $\alpha$-softmax function and the other applying $\alpha$-quasimax function. It is noted that models of the proposed approach consistently outperform WSIS models at large margins, indicating that it is advantageous to introduce bounding-box regression in the proposed approach for CDR measurement. 

To evaluate the advantages of including weakly supervised image segmentation in the proposed approach, the results of the RetinaNet model are also listed in Table \ref{table:main}. As noted, the RetinaNet model obtains CDR error of 0.0468 and $F_1$ score of 0.903, worse than those from the proposed approach. These results imply that it is beneficial to include weakly supervised image segmentation in the proposed approach for CDR measurement.

\subsubsection{Auxiliary results: performance of OC and OD segmentation}

In Table \ref{table:main_seg} we provide the results of the proposed approach on the testing set for OC and OD segmentation, in which the performance of image segmentation was measured by dice coefficient. For comparison, dice coefficients of the FSIS model are also given in Table \ref{table:main_seg}. As can be seen, both models of the proposed approach get dice coefficients close to the FSIS model. Particularly, the proposed approach using $\alpha$-softmax function has dice coefficient of 0.882 for OC segmentation, close to 0.891 obtained by the FSIS model; it gets dice coefficient of 0.950 for OD segmentation, close to 0.969 from the FSIS model.

\begin{table}
\caption{Performance comparison of the proposed CDRNet and different existing methods in OC and OD segmentation. The methods as follows are considered for comparison: 1) FSIS, 2) WSIS, and 3) RetinaNet.}
\centering
\begin{tabular}{cccc}
\hline
\hline
 & OC & OD & Avg. \\
\hline
CDRNet ($\alpha$-softmax) & 0.882 & 0.950 & 0.916 \\
CDRNet ($\alpha$-quasimax) & 0.881 & 0.950 & 0.915 \\
FSIS & 0.891 & 0.969 & 0.930 \\
WSIS ($\alpha$-softmax) & 0.870 & 0.955 & 0.913 \\
WSIS ($\alpha$-quasimax) & 0.873 & 0.955 & 0.914 \\
RetinaNet & - & - & - \\
\hline
\hline
\end{tabular}
\label{table:main_seg}
\end{table}

The dice coefficients of WSIS models are also given in Table \ref{table:main_seg} for OC and OD segmentation. Compared with the WSIS models, both models of the proposed approach get higher dice coefficients for OC segmentation, but lower dice coefficients for OD segmentation. As a performance summary, the average dice coefficients of OC and OD segmentation are also listed in the last column of Table \ref{table:main_seg}. It can be observed that both models of the proposed approach get higher average dice coefficients (0.916 for the proposed approach using $\alpha$-softmax function and 0.915 for the proposed approach using $\alpha$-quasimax function) when compared with WSIS models (0.913 for the WSIS model using $\alpha$-softmax function and 0.914 for the WSIS model using $\alpha$-quasimax function). These results indicate that introducing bounding-box regression in the proposed approach is also helpful for OC and OD segmentation. 

Considering the performance of CDR measurement and OC and OD segmentation together, from Tables \ref{table:main} and \ref{table:main_seg}, it can be noted that the proposed approach using $\alpha$-softmax function achieves better performance than that using $\alpha$-quasimax function. Therefore, in the following analysis, we will consider the proposed CDRNet using $\alpha$-softmax function and refer it as the proposed approach if no specific explanations are provided.

\subsection{Performance comparison of the proposed CDRNet and individual graders}
\subsubsection{Main results: performance of CDR measurement}
To compare the performance of the proposed approach with those of individual graders in CDR measurement, in Table \ref{table:cdrnet_readerstudy_cdr} we provide CDR errors of the proposed approach and four graders in a pairwise fashion. Owing to the symmetry of the CDR error, the entries in the lower triangular portion are omitted in Table \ref{table:cdrnet_readerstudy_cdr}. As can be seen, the CDR error of the proposed approach is lowest when the readings of $G_3$ are treated as ground truth, and highest when the readings from $G_1$ are used as ground truth. More importantly, the proposed approach achieves average CDR error of 0.0430, which is much lower than those by individual graders (i.e. 0.0546 for $G_1$, 0.0507 for $G_2$, 0.0476 for $G_3$, and 0.0513 for $G_4$). These results suggest that the proposed approach outperforms individual graders in CDR measurement. The similar conclusion can be made when $F_1$ score is considered, we omit here for conciseness.

\begin{table}
\caption{Performance comparison of the proposed approach and different graders in CDR measurement.}
\centering
\setlength{\tabcolsep}{1.5pt}
\begin{tabular}{ccccccc}
\hline
\hline
& CDRNet & $G_1$ & $G_2$ & $G_3$ & $G_4$ & Average \\
\hline
CDRNet & 0 & 0.0486 & 0.0418 & 0.0399 & 0.0415 & \textbf{0.0430} \\
$G_1$ & - & 0 & 0.0599 & 0.0530 & 0.0568 & 0.0546 \\
$G_2$ & - & - & 0 & 0.0459 & 0.0554 & 0.0507 \\
$G_3$ & - & - & - & 0 & 0.0516 & 0.0476 \\
$G_4$ & - & - & - & - & 0 & 0.0513 \\
\hline
\hline
\end{tabular}
\label{table:cdrnet_readerstudy_cdr}
\end{table}

\subsubsection{Auxiliary results: performance of OC and OD segmentation}

In addition, in Table \ref{table:cdrnet_readerstudy_dice} we also show the results of the proposed approach and four graders in a pairwise fashion for OC and OD segmentation, in which the average dice coefficient of OC and OD segmentation is considered. It can be observed that the proposed approach gets average dice coefficients in a small range of 0.923 and 0.930 when the readings from the four graders are treated as ground truth. Moreover, the average of the average dice coefficients is 0.925 for the proposed approach, close to those from individual graders (i.e. 0.925 for $G_1$, 0.923 for $G_2$, 0.928 for $G_3$, and 0.923 for $G_4$). These results indicate that the proposed approach has segmentation performance similar to individual graders.

\begin{table}
\caption{Performance comparison of the proposed approach and different graders in OC and OD segmentation, in which the average dice coefficient of OC and OD segmentation is considered.}
\centering
\setlength{\tabcolsep}{2pt}
\begin{tabular}{ccccccc}
\hline
\hline
& CDRNet & $G_1$ & $G_2$ & $G_3$ & $G_4$ & Average \\
\hline
CDRNet & 1 & 0.923 & 0.923 & 0.930 & 0.924 & 0.925 \\
$G_1$ & - & 1 & 0.923 & 0.932 & 0.923 & 0.925 \\
$G_2$ & - & - & 1 & 0.925 & 0.920 & 0.923 \\
$G_3$ & - & - & - & 1 & 0.926 & 0.928 \\
$G_4$ & - & - & - & - & 1 & 0.923 \\
\hline
\hline
\end{tabular}
\label{table:cdrnet_readerstudy_dice}
\end{table}

\subsection{Robustness of the proposed CDRNet on different hyperparameters}

\subsubsection{Sensitivity of the proposed approach on parameter $\sigma$ in smooth $L_1$ loss}

To evaluate the sensitivity of the proposed approach on parameter $\sigma$ in smooth $L_1$ loss, Table \ref{table:cdrnet_sigma} shows the results of the proposed approach in CDR measurement for different $\sigma$'s. As observed, compared with the best performance at $\sigma=6$ (CDR error = 0.0458, $F_1$ score = 0.917), there is only slight performance decreasing for CDR error and $F_1$ score when $\sigma < 6$ and $\sigma > 6$. These results suggest that the proposed approach is robust to $\sigma$ in smooth $L_1$ loss for CDR measurement.

\begin{table}
\caption{Performance comparison of the proposed approach for different $\sigma$'s in smooth $L_1$ loss.}
\centering
\begin{tabular}{ccccc}
\hline
\hline
 & CDR error & $F_1$ score & Dice OC & Dice OD \\
\hline
$\sigma=3$ & 0.0466 & 0.915 & 0.880 & 0.950 \\
$\sigma=4$ & 0.0480 & 0.913 & 0.879 & 0.949 \\
$\sigma=5$ & 0.0471 & 0.907 & 0.880 & 0.949 \\
$\sigma=6$ & 0.0458 & 0.917 & 0.882 & 0.950 \\
$\sigma=7$ & 0.0461 & 0.913 & 0.881 & 0.951 \\
$\sigma=8$ & 0.0469 & 0.906 & 0.880 & 0.950 \\
\hline
\hline
\end{tabular}
\label{table:cdrnet_sigma}
\end{table}

Moreover, the results of the proposed approach in OC and OD segmentation are listed in Table \ref{table:cdrnet_sigma} for different $\sigma$'s in smooth $L_1$ loss. It is observed that the dice coefficients of both OC and OD classes are very close for all $\sigma$'s, indicating that the proposed approach is insensitive to $\sigma$ for image segmentation.

\subsubsection{Sensitivity of the proposed approach on threshold $T$ in positive sample selection}

To demonstrate the robustness of the proposed approach on threshold $T$ in positive sample selection, in Table \ref{table:cdrnet_T}, CDR errors of the proposed approach are given for different $T$'s, in which $T=0$ indicates that no positive sample selection was applied. As can be seen, at $T=0$, CDR error (0.0476) is higher than the optimal setting ($T=0.6$) and $F_1$ score (0.909) is lower than the optimal setting, demonstrating the effectiveness of the proposed positive sample selection. Moreover, compared to the optimal setting of $T=0.6$, both $T=0.5$ and $T=0.7$ get slightly decreased performance (CDR error = 0.0480, $F_1$ score = 0.908 for $T=0.5$, and CDR error = 0.0480, $F_1$ score = 0.904 for $T=0.7$).

\begin{table}
\caption{Performance comparison of the proposed approach for different threshold $T$'s in positive sample selection.}
\centering
\begin{tabular}{ccccc}
\hline
\hline
 & CDR error & $F_1$ score & Dice OC & Dice OD \\
\hline
$T=0$ & 0.0476 & 0.909 & 0.882 & 0.948 \\
$T=0.5$ & 0.0467 & 0.908 & 0.882 & 0.950 \\
$T=0.6$ & 0.0458 & 0.917 & 0.882 & 0.950 \\
$T=0.7$ & 0.0480 & 0.904 & 0.878 & 0.951 \\
\hline
\hline
\end{tabular}
\label{table:cdrnet_T}
\end{table}

In addition, the results of the proposed approach in OC and OD segmentation are listed in Table \ref{table:cdrnet_T} for different $T$'s. The dice coefficients for both OC and OD classes are very close in different $T$'s, indicating that the image segmentation predictions are robust to sample selection threshold $T$ in the proposed approach.

\subsection{Additional results}

\subsubsection{Model and inference complexity} 

In Table \ref{table:params_time}, the results of both model and inference complexity of different methods are reported. For each method, the number of trainable parameters and FLOPS are reported for model complexity, the inference time and the throughput in FPS are used for inference complexity. As noted, the proposed approach takes about 45.2 ms/image for inference and has a throughput of 22 FPS on a GPU.

\begin{table}
\caption{Comparison of the proposed approach and different existing methods in the model and inference complexity, including the number of parameters, FLOPS, inference time (ms/image), and throughput in FPS.}
\centering
\begin{tabular}{ccccc}
\hline
\hline
methods & \# parameters & FLOPS & inference time & throughput \\
\hline
CDRNet & 26,360,666 & 955 G & 45.2 & 22 \\
FSIS & 23,170,386 & 170 G & 15.2 & 66 \\
WSIS & 23,170,386 & 170 G & 15.2 & 66 \\
RetinaNet & 30,136,264 & 144 G & 14.9 & 67 \\
\hline
\hline
\end{tabular}
\label{table:params_time}
\end{table}

\subsubsection{Result visualization of the proposed CDRNet}

\begin{figure}[htbp] 
	\centering
	\setlength{\tabcolsep}{2pt}
	\begin{tabular}{ccc}
	\includegraphics[trim=0in 0in 0in 0in,clip,width=1.5in]{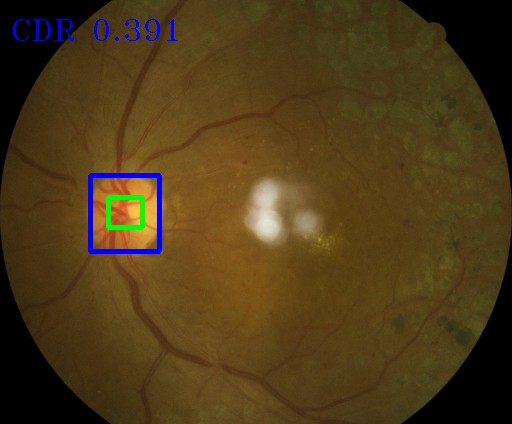} &
	\includegraphics[trim=0in 0in 0in 0in,clip,width=1.5in]{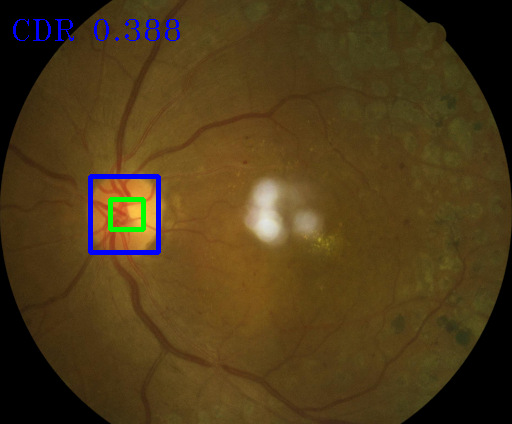} &
	\includegraphics[trim=0in 0in 0in 0in,clip,width=1.5in]{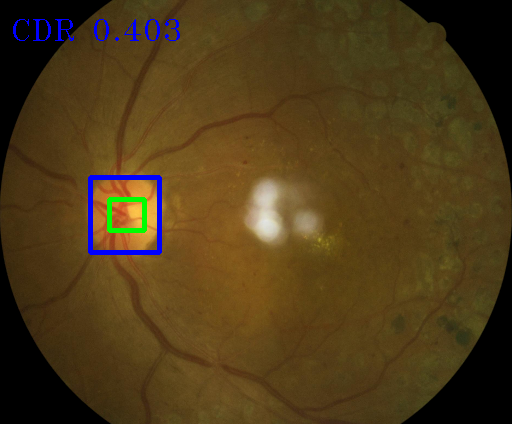} \\ 
	\includegraphics[trim=0in 0in 0in 0in,clip,width=1.5in]{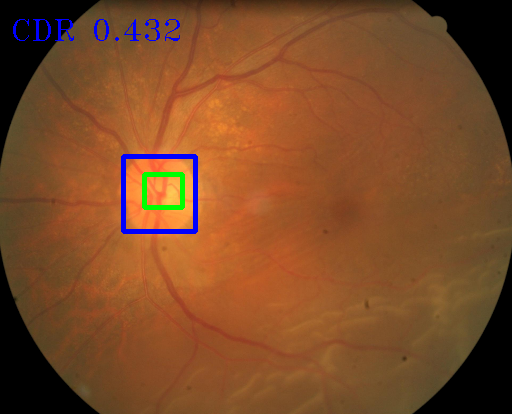} &
	\includegraphics[trim=0in 0in 0in 0in,clip,width=1.5in]{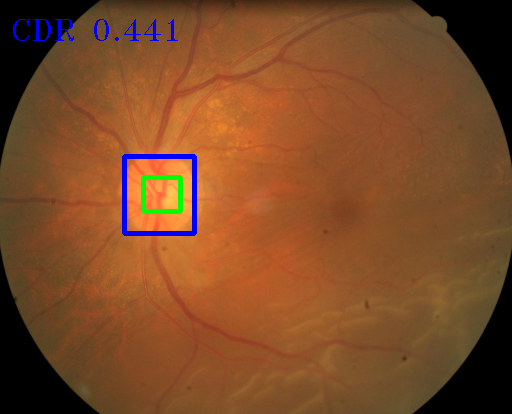} &
	\includegraphics[trim=0in 0in 0in 0in,clip,width=1.5in]{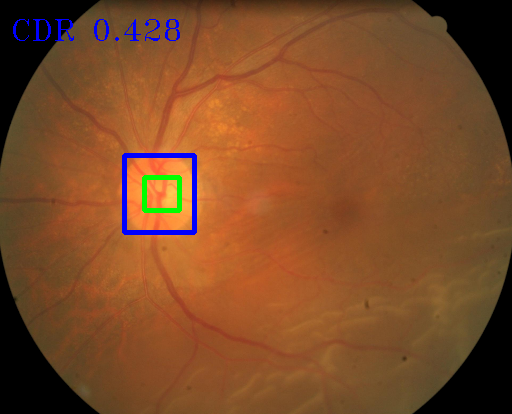} \\ 
	\includegraphics[trim=0in 0in 0in 0in,clip,width=1.5in]{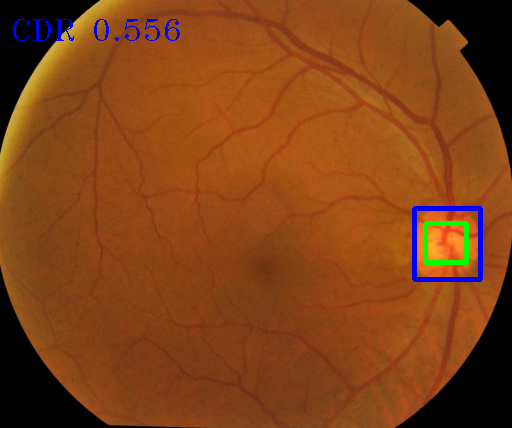} &
	\includegraphics[trim=0in 0in 0in 0in,clip,width=1.5in]{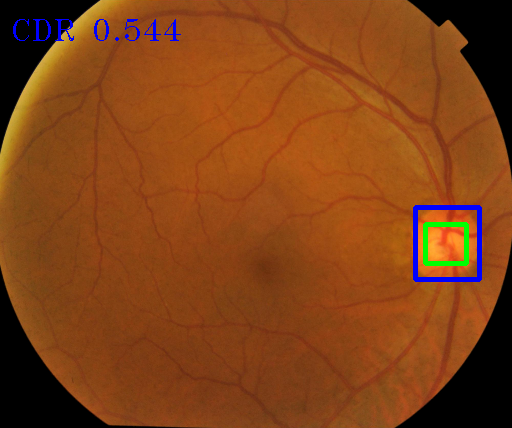} &
	\includegraphics[trim=0in 0in 0in 0in,clip,width=1.5in]{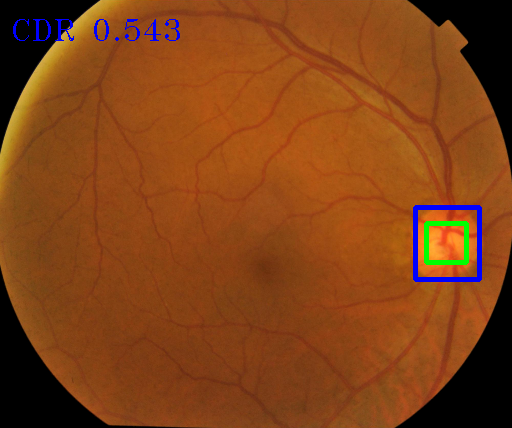} \\ 
	\includegraphics[trim=0in 0in 0in 0in,clip,width=1.5in]{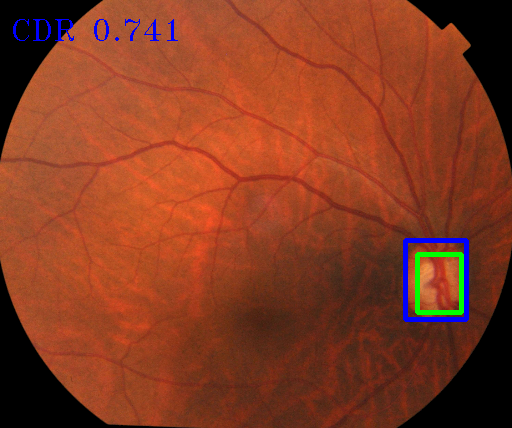} &
	\includegraphics[trim=0in 0in 0in 0in,clip,width=1.5in]{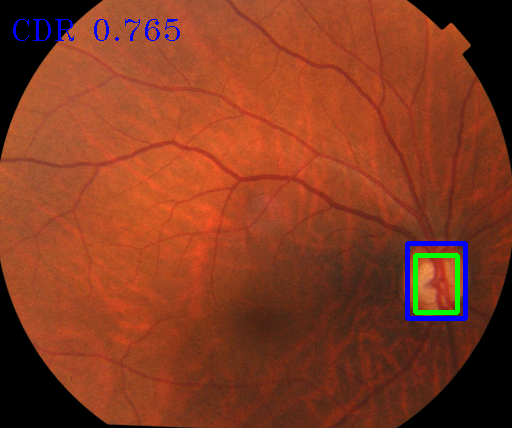} &
	\includegraphics[trim=0in 0in 0in 0in,clip,width=1.5in]{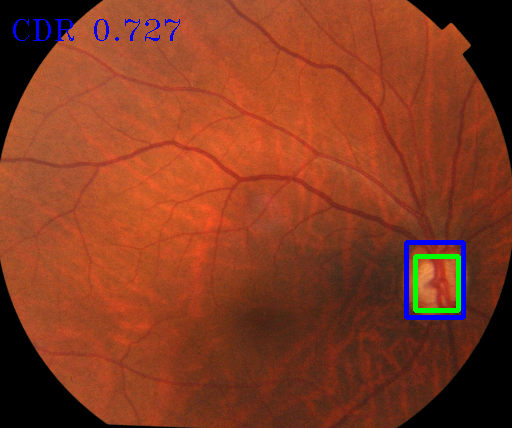} \\ 
	(a) & (b) & (c) \\
	\end{tabular}
	\caption{Demonstration of (a) ground-truth bounding boxes, (b) predicted bounding boxes for the proposed CDRNet using $\alpha$-softmax, and (c) predicted bounding boxes for the proposed CDRNet using $\alpha$-quasimax in the testing set. In these plots, the bounding boxes of OC and OD are denoted by green and blue colors, respectively. For clarity, the corresponding CDR values are shown in the images as well.}
	\label{fig:det_demonstration}
\end{figure}

\begin{figure}[htbp] 
	\centering
	\setlength{\tabcolsep}{2pt}
	\begin{tabular}{ccc}
	\includegraphics[trim=0in 0in 0in 0in,clip,width=1.5in]{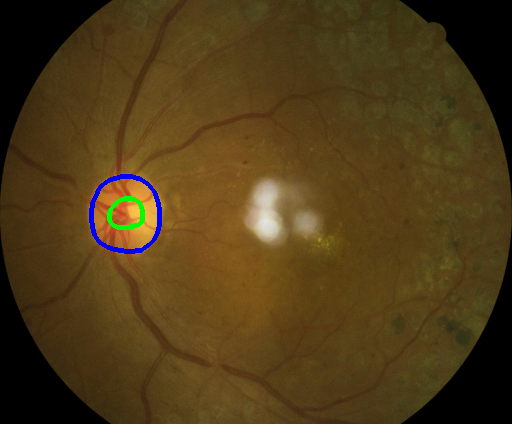} &
	\includegraphics[trim=0in 0in 0in 0in,clip,width=1.5in]{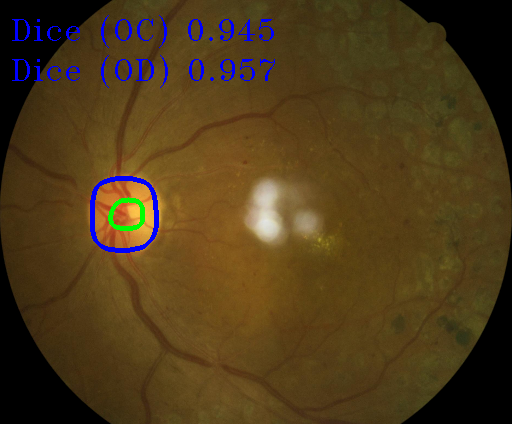} &
	\includegraphics[trim=0in 0in 0in 0in,clip,width=1.5in]{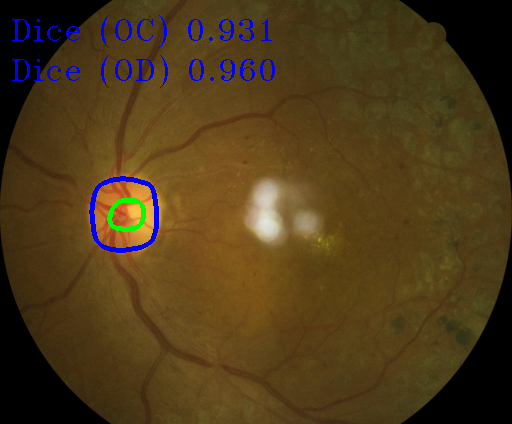} \\ 
	\includegraphics[trim=0in 0in 0in 0in,clip,width=1.5in]{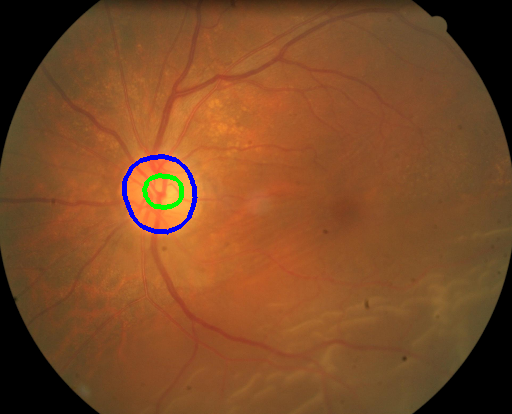} &
	\includegraphics[trim=0in 0in 0in 0in,clip,width=1.5in]{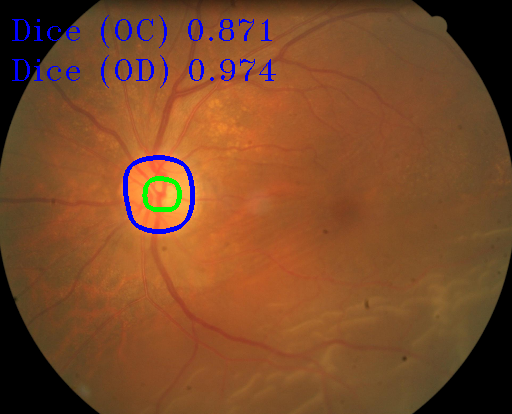} &
	\includegraphics[trim=0in 0in 0in 0in,clip,width=1.5in]{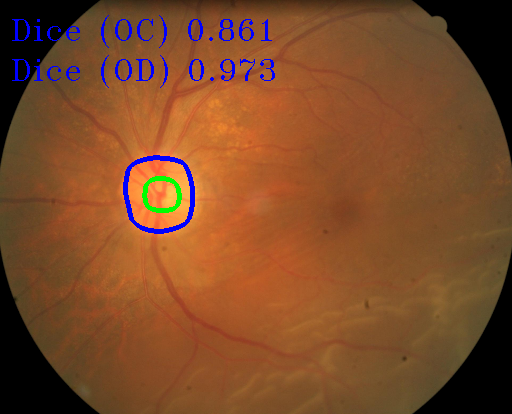} \\ 
	\includegraphics[trim=0in 0in 0in 0in,clip,width=1.5in]{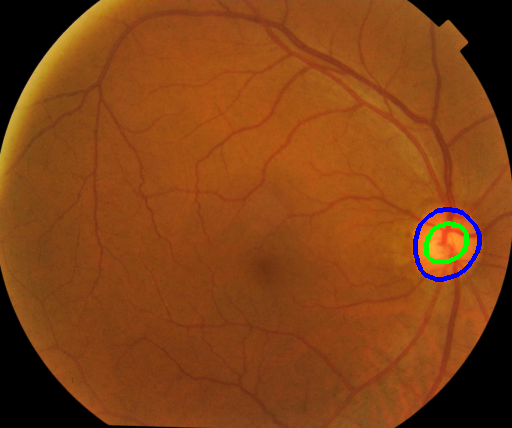} &
	\includegraphics[trim=0in 0in 0in 0in,clip,width=1.5in]{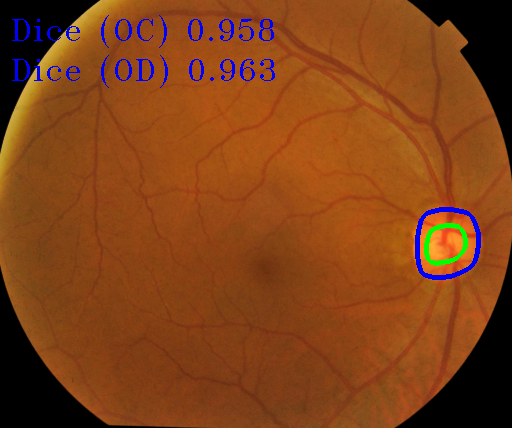} &
	\includegraphics[trim=0in 0in 0in 0in,clip,width=1.5in]{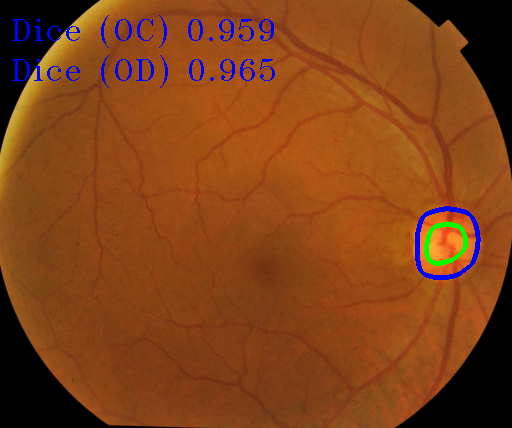} \\ 
	\includegraphics[trim=0in 0in 0in 0in,clip,width=1.5in]{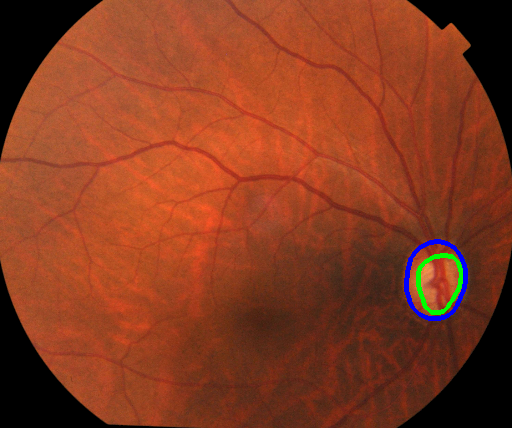} &
	\includegraphics[trim=0in 0in 0in 0in,clip,width=1.5in]{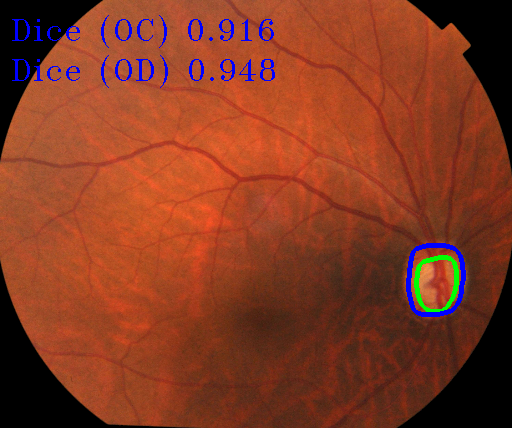} &
	\includegraphics[trim=0in 0in 0in 0in,clip,width=1.5in]{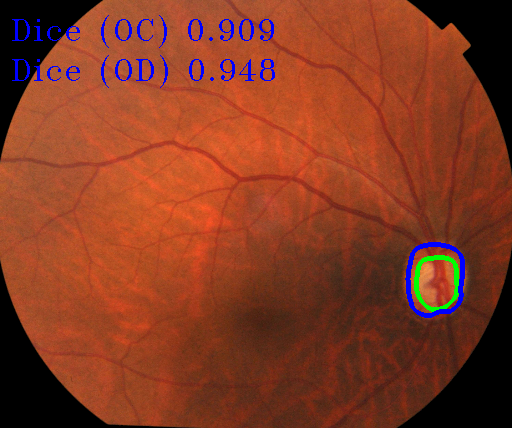} \\ 
	(a) & (b) & (c) \\
	\end{tabular}
	\caption{Demonstration of (a) ground-truth boundaries of OC and OD masks, (b) predicted boundaries of OC and OD masks for the proposed approach using $\alpha$-softmax (middle), and (c) predicted boundaries of OC and OD masks for the proposed approach using $\alpha$-quasimax in the testing set. In these plots, boundaries of OC and OD masks are denoted by green and blue colors, respectively. For clarity, the dice coefficients of the predicted OC and OD masks are shown in the images as well.}
	\label{fig:seg_demonstration}
\end{figure}

Finally, to visually demonstrate the performance of the proposed CDRNet,
bounding-box predictions and their corresponding CDR values are shown for $\alpha$-softmax setting in Figure \ref{fig:det_demonstration} (b) and for $\alpha$-quasimax setting in Figure \ref{fig:det_demonstration}(c). For reference, the ground truth of bounding boxes and their corresponding CDR values are shown in Figure \ref{fig:det_demonstration}(a) as well. It can be seen that the proposed method achieves good bounding-box predictions, resulting in accurate CDR values. Similarly, the quantitative results of OC and OD segmentation are given in Figure \ref{fig:seg_demonstration}.

\section{Discussions}
This study developed CDRNet, a two-task network for CDR measurement using tight bounding box supervision. As demonstrated in Section \ref{sec:results}, compared with the existing methods in the literature, the proposed CDRNet has advantages in several aspects as follows: 1) efficiency in data annotation, 2) superior performance, and 3) richness in output. For efficiency in data annotation, the proposed network considers weak annotations as supervision (i.e. tight bounding-box annotations) while achieving better performance in CDR measurement and similar performance in image segmentation when compared with the FSIS method using full supervision (i.e. pixel-wise annotations). For superior performance, among the methods using weak supervision based on the tight bounding-box annotations, the proposed network gets improved performance in both CDR measurement and image segmentation when compared with the WSIS method; it also achieves better performance in CDR measurement when compared with RetinaNet. For richness in output, the proposed network directly outputs both segmentation results and bounding-box regression results while the existing methods for comparison outputs only one of them.

As noted in the introduction, the reason that the proposed approach is accurate in CDR measurement is because of its superior performance in object size measurement. To demonstrate this, in Table \ref{table:size_error} we show the mean absolute difference (MAD) of the vertical diameter of the object (i.e. OC and OD) between the bounding box label and its corresponding prediction from the proposed approach. For comparison, WSIS and RetinaNet, both of which employed tight bounding box supervision, are also considered in Table \ref{table:size_error}. As can be seen, the proposed approach get MAD value of 3.207 for OC and 2.067 for OD, which are much lower than those from WSIS models. Moreover, compared with MAD value of 3.280 for OC and 2.162 for OD from RetinaNet, the proposed approach yields a 2.23\% reduction in MAD for OC and a 4.39\% reduction in MAD for OD. These results confirm the superior performance of the proposed approach in object size measurement. 

\begin{table}
\caption{The performance comparison of object size measurement for the proposed approach and different existing methods (i.e. WSIS and RetinaNet). The object size under consideration are the vertical diameters of OC and OD.}
\centering
\begin{tabular}{ccc}
\hline
\hline
 & OC & OD \\
\hline
CDRNet & 3.207 & 2.067 \\
WSIS ($\alpha$-softmax) & 6.143 & 4.474 \\
WSIS ($\alpha$-quasimax) & 6.068 & 4.824 \\
RetinaNet & 3.280 & 2.162 \\
\hline
\hline
\end{tabular}
\label{table:size_error}
\end{table}

Finally, it has to be noted that only the feature maps with same dimension as input image are considered for bounding-box regression in this study. It is a good design choice for bounding-box regression of OC and OD considering the fact that the sizes of objects in both OC and OD classes vary in a small range among different images. However, in the application when the sizes of objects in a class vary hugely, the proposed approach might yield decreased performance and such techniques as feature pyramid network should be considered for the bounding-box regression.

\section{Conclusion}
This study developed a two-task network for CDR measurement using tight bounding box supervision, one for weakly supervised image segmentation, and the other for bounding-box regression. The weakly supervised image segmentation task is based on generalized MIL formulation and smooth maximum approximation developed in our previous study. The bounding-box regression task outputs class-specific bounding box prediction in a single scale at the original image resolution. A class-specific bounding-box normalizer and an eIoU were proposed for accurate bounding box predictions. The experimental results demonstrate that the proposed network outperforms the state-of-the-art performance for CDR measurement and gets performance close to the state-of-the-art for OC and OD segmentation. Compared with the individual graders, the proposed approach achieves better performance for CDR measurement and similar performance for OC and OD segmentation. One limitation of the proposed approach is the measurement performance might be decreased if the sizes of the objects in a class vary hugely. For such applications, it would be interesting to improve the bounding-box regression by incorporating such techniques as feature pyramid network in the future.

\bibliographystyle{unsrt}
\bibliography{reference}  






\end{document}